\documentclass[conference,compsoc]{IEEEtran}

\usepackage{cite}

\ifCLASSINFOpdf
   \usepackage[pdftex]{graphicx}
\else
   \usepackage[dvips]{graphicx}
\fi
\usepackage[cmex10]{amsmath}
\usepackage{algorithmic}
\usepackage{algorithm}

\usepackage{array}
\usepackage{mdwmath}
\usepackage{mdwtab}
\usepackage{eqparbox}

\usepackage[font=footnotesize]{subfig}

\usepackage{url}

\begin{document}
%
\title{Mobile Camera Array Calibration for Light Field Acquisition}
\author{\IEEEauthorblockN{Yichao Xu, Kazuki Maeno, Hajime Nagahara, Rin-ichiro Taniguchi}
\IEEEauthorblockA{Graduate School of Information Science and Electrical Engineering, Kyushu University\\
744 Motooka, Nishi-ku, Fukuoka 819-0395, Japan\\
\{xu,maeno,nagahara,rin\}@limu.ait.kyushu-u.ac.jp}}
\maketitle

\begin{abstract}
The light field camera is useful for computer graphics and vision applications. Calibration is an essential step for these applications. After calibration, we can rectify the captured image by using the calibrated camera parameters. However, the large camera array calibration method, which assumes that all cameras are on the same plane, ignores the orientation and intrinsic parameters. The multi-camera calibration technique usually assumes that the working volume and viewpoints are fixed. In this paper, we describe a calibration algorithm suitable for a mobile camera array based light field acquisition system. The algorithm performs in Zhang's style by moving a checkerboard, and computes the initial parameters in closed form. Global optimization is then applied to refine all the parameters simultaneously. Our implementation is rather flexible in that users can assign the number of viewpoints and refinement of intrinsic parameters is optional. Experiments on both simulated data and real data acquired by a commercial product show that our method yields good results. Digital refocusing application shows the calibrated light field can well focus to the target object we desired.

\end{abstract}

\IEEEpeerreviewmaketitle

\section{Introduction}
The light field camera was designed as a device to record the distribution of light rays in space. It can capture a 4D light field, which includes both positional and angular information. Because the light field camera can obtain a 4D profile of the light rays, it can produce effects well beyond the capabilities of regular cameras. The light field camera is useful for both computer graphics and computer vision applications.
\begin{itemize}
  \item \textbf{Image based rendering} \cite{Levoy96-LFR} can be performed if the light field has been captured. An image from any viewpoint can be computed by re-sampling of the acquired light rays.
  \item \textbf{Synthetic aperture photography} \cite{synthetic06} also uses the light field. Images can be refocused to any focal plane in a scene by re-projection of the acquired light rays onto the target plane. When the synthetic aperture is large enough, occluding objects in front of the focal plane are blurred to the extent that they effectively disappear, and the occluded objects can then be seen.
  \item \textbf{Scene geometry reconstruction} from multiple views has been an active area of research in computer vision \cite{multi-view06} for some time. The light field camera can capture multiple viewpoint images simultaneously, so that the 3D geometry can be recovered efficiently from a single shot. By using the light field distortion, even transparent surfaces can be reconstructed using a single camera \cite{WetzsteinRHR11}.
\end{itemize}
Light field camera calibration is an essential step for both the computer graphics and computer vision applications. After calibration, we can not only recover the camera parameters, but can also recover the positions and orientations of the light field for each of the viewpoints. We can then use the calibrated parameters to rectify the captured light field.

In the early days of the method, light fields were captured by a camera moving on a controlled gantry \cite{Levoy96-LFR}. A single camera is easy to calibrate by classical calibration techniques. The positions and orientations of each viewpoint can be determined from the gantry position. The large camera array \cite{camArr02} is also a powerful piece of equipment for light field acquisition. In applications, the large camera array usually captures scenes that are far away from the camera array, and the field of view is very narrow so that the images only suffer small lens distortion. The cameras in the array are effectively on a plane, and thus the displacement on the principal axis can be ignored for long-range applications. Therefore, in most cases, large camera array calibration simply requires recovery of the 2D position of the cameras and the 2D pixel coordinates on the reference plane.

Obtaining the light field by using either a gantry or a large camera array originally required large-scale equipment which was both expensive and difficult to operate. In recent years, however, commercial cameras for light field capture have become smaller and less expensive. Representative products include the mobile camera array\cite{profusion25}, and the plenoptic camera \cite{lytro}, which consists of a micro-lens array between the sensor and the main lens. The manufacturers do not usually provide calibration applications for these commercial products, and these products are quite different to classical light field acquisition equipment. When using a commercial product for our applications, we must develop a calibration method that is suitable for the light field camera. In this paper, we focus on mobile camera array based light field camera calibration.

When we apply a mobile camera array to a vision task such as object recognition or scene geometry reconstruction, it requires accurate calibration. Large camera arrays are usually applied to long-range visualization applications, which do not require precise calibration. The calibration method used for a large camera array is therefore not suitable for the mobile camera array. Also, the calibration methods for used multi-camera calibration usually assume that the working volume is fixed and that the cameras are static, while the mobile camera array can be moved freely to any position. In this paper, we propose a method for calibration of the mobile camera array, which includes the following contributions:
\begin{enumerate}
  \item The position-angular representation of the light field is defined first, and then the model for the mobile camera array is given.
  \item A global optimization-based algorithm is proposed for calibration of the mobile camera array. The algorithm is flexible, so that the user can decide on the estimation parameters and the number of viewpoints required.
  \item We apply the proposed method to a commercial light field camera product, and compare its performance with that of other methods.
\end{enumerate}

\section{Related Work}
To acquire the camera parameters with high accuracy, a great deal of work has been done on camera calibration over the last few decades. Based on the number of cameras in the system, we can classify the calibration methods into three categories: calibration for single cameras, multiple cameras and large camera arrays.

\textbf{Single camera calibration:}
Classical camera calibration is performed by observation of a 3D reference object with a known Euclidean geometry\cite{Tsai87a}. This kind of approach requires specialized and expensive equipment with an elaborate setup. To avoid these disadvantages, a flexible technique for single camera calibration was proposed by Z. Zhang\cite{zhangcalib}, which only required the camera to observe a planar pattern shown at a minimum of two different orientations. The pattern can simply be printed using a laser printer and then attached to a "reasonable" planar surface (e.g., a hard book cover). Either the camera or the planar pattern can be moved by hand. The specific motion need not be known. This technique is very practical and robust for a single camera, but it is not suitable for a light field camera. The rigid transformations between any pair of viewpoints, which could be determined through any captured frame, should be invariant irrespective of the frame through which they are computed (see Fig. \ref{camGeo}). Unfortunately, these transformations are inconsistent when each viewpoint is calibrated independently. This inconsistency would cause inaccurate estimation of the relative displacements between the viewpoints, potentially leading to serious problems if used for multi-camera systems and light field cameras.

\textbf{Multi-camera calibration:}
Because multi-camera systems are becoming less expensive and more useful, there are increasing requirements for multi-camera system calibration. New techniques are being developed to deal with multi-camera systems. Ueshiba et al.\cite{multi-camCalib} proposed a method that also uses a planar pattern. The method uses homography matrices between the camera's image and the planar pattern to compose a measurement matrix with an unknown scale and then factorizes this measurement matrix into the camera and plane parameters. However, this method needs additional information from the "relative homography" between the different camera images and the different frames to overcome the unknown scale, and the lens distortion is not considered in this method.
A more convenient method for multi-camera calibration was proposed by Svoboda et al.\cite{Svoboda-Presence2005}. This method also used a factorization approach, but instead of the planar pattern, the calibration object is simply a freely moving bright spot. However, this method is specific to static multi-camera systems, where the system's working volume is fixed, while the global coordinates of a light field camera can be moving freely.

\textbf{Large camera array calibration:}
Large camera arrays are built for light field acquisition, and require calibration of the cameras used to acquire the light field. Vaish et al.\cite{camArrCal} proposed a method using plane plus parallax to calibrate large camera arrays. The camera positions can be recovered when the cameras lie on a plane that is parallel to the reference plane. To recover the camera positions, the method measures the parallax of a single scene point that is not on this reference plane. Once the relative camera positions have been acquired, the light field can then be represented in a two-plane parametrization. This method, however, assumes that all cameras are on the same plane and must calculate the projection to a reference plane in advance. Only a few applications that do not require accurate parameters can use this calibration technique.

\section{Light Field Camera Modeling}

\subsection{Projection model}
\begin{figure}[!t]
\centering
\includegraphics[width=3in]{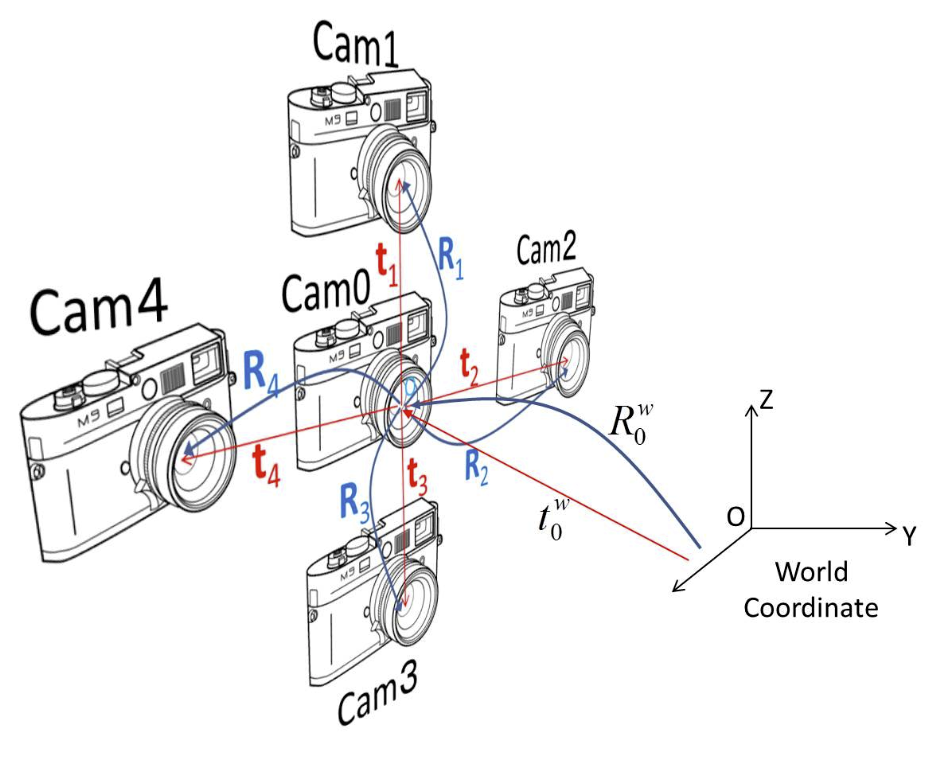}
\caption{Camera Array Geometry}
\label{camGeo}
\end{figure}
Let us consider a system that contains $N$ cameras, so that the light field camera system can simultaneously capture images from $N$ viewpoints. Each viewpoint independently records its own 2D image as $L(x_p,y_p)$ from its position. The projection of a 3D point $\mathbf{M}=[X,Y,Z]^T$ to a 2D point $\mathbf{m}=[x_p,y_p]^T$ on the image plane of the $i$-th viewpoint is given by
\begin{equation}
\label{3D2img}
\left[
  \begin{array}{c}
	\mathbf{m}\\
    1
  \end{array}
\right]
\simeq \mathbf{A}_i[\mathbf{R}_i^{w} \quad \mathbf{t}_i^{w}]
\left[
  \begin{array}{c}
	\mathbf{M}\\
    1
  \end{array}
\right]
\end{equation}
where $[\mathbf{R}_i^{w} \quad \mathbf{t}_i^{w}]$ represents the extrinsic parameters and consists of the rotation matrix and the translation vector, and $\mathbf{A}_i$ is called the intrinsic matrix, which is given by
\begin{equation}
\label{inMat}
\mathbf{A}_i=
\left[
  \begin{array}{ccc}
	\alpha _i & \gamma _i & u_{0i}\\
	0      & \beta _i  & v_{0i}\\
    0      & 0      & 1
  \end{array}
\right]
\end{equation}
and contains the coordinates of the principal point $(u_{0i},v_{0i})$, the focal length $\alpha _i$, $\beta _i$, and the skew of the two image axes $\gamma _i$.

We assume that the global coordinate system is fixed to the first viewpoint. The relative position and orientation of the $i$-th viewpoint can be represented by a relative translation vector $\mathbf{t}_{i}$ and a relative rotation matrix $\mathbf{R}_{i}$, respectively (Fig. \ref {camGeo}). Obviously, the relative position and orientation of the first viewpoint are $\mathbf{t}_0=\mathbf{0}$ and $\mathbf{R}_0=\mathbf{I}$.
\begin{figure}[!t]
\centering
\includegraphics[width=3in]{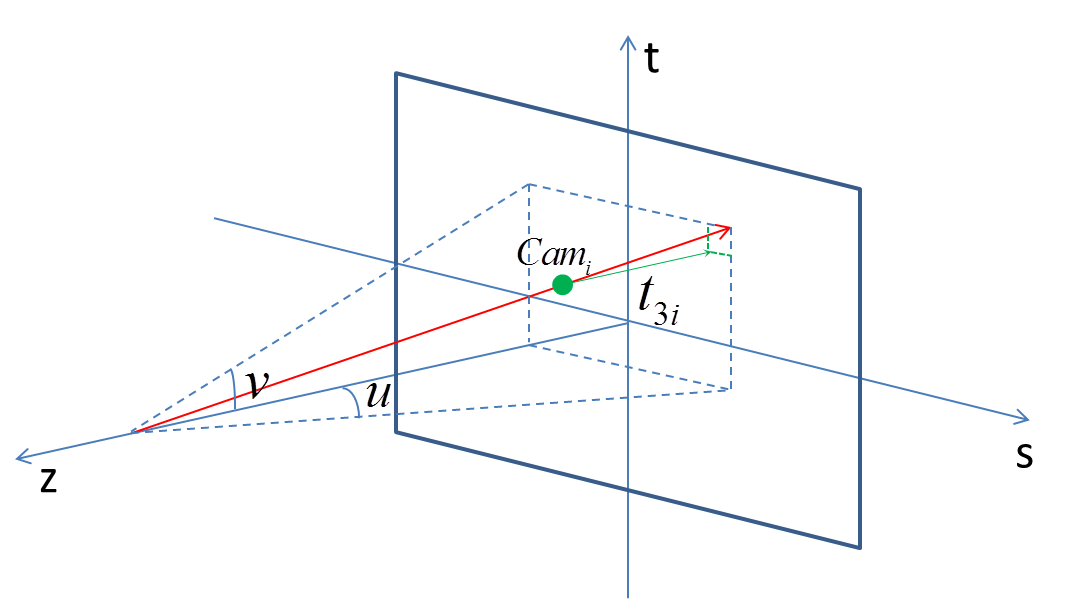}
\caption{Definition of light field representation $L(s,t,u,v)$}
\label{LF_Def}
\end{figure}

Once we obtain the parameters of $[\mathbf{R}_i \quad \mathbf{t}_i]$ and $\mathbf{A}_i$, we can then represent the light field image as $L(s,t,u,v)$ (see Fig. \ref{LF_Def}) by projecting these images onto the parallel image coordinates prescribed by slant $(u,v)$. The projection from the raw images $L(x_p,y_p)$ to the slant of the light field $L(u,v)$ can be calculated from:
\begin{equation}
\label{pts2slants}
\left[
  \begin{array}{c}
	u\\
    v\\
    1
  \end{array}
\right]
= ^\infty\mathbf{H}_i\cdot\mathbf{A}_i^{-1}
\left[
  \begin{array}{c}
	x_p\\
    y_p\\
    1
  \end{array}
\right]
\end{equation}
where $^\infty\mathbf{H}_i$ is a matrix that describes the infinite homography between two views:
\begin{equation}
\label{rel2hom}
^\infty\mathbf{H}_i=\mathbf{R}_0\cdot\mathbf{R}_{i}^{-1}=\mathbf{R}_{i}^{-1}
\end{equation}
If $t_{3i}$ of the translation vector $\mathbf{t}_i$=$[t_1,t_2,t_3]_i^T$ is very small, we can ignore the translation along the principal axis, and then the coordinates are $(s,t)$=$(t_1,t_2)$. Otherwise, if $t_{3i}$ is large enough to affect the coordinates $(s,t)$, we should calculate the new $(s,t)$ coordinates as follows:
\begin{equation}
\label{t2s}
\left[
  \begin{array}{c}
	s\\
    t
  \end{array}
\right]
= t_{3i}\cdot
\left[
  \begin{array}{c}
	u\\
    v
  \end{array}
\right]
+
\left[
  \begin{array}{c}
	t_{1i}\\
    t_{2i}
  \end{array}
\right]
\end{equation}
When computing the projection of each viewpoint, we require the absolute position and orientation of these viewpoints in the global coordinate system. These parameters can be computed from:
\begin{equation}
\label{rel2ext}
\begin{cases}
\mathbf{R}_i^{w}=\mathbf{R}_i\cdot\mathbf{R}_{0}^{w}\\
\mathbf{t}_i^{w}=\mathbf{R}_i\cdot\mathbf{t}_{0}^{w}+\mathbf{t}_i
\end{cases}
\end{equation}
We have now constructed the light field camera model, and the unknown parameters can be solved by using the linear closed form and refined by nonlinear optimization in the following steps.

\subsection{Camera lens distortion}
Camera lenses usually suffer from distortion, particularly radial distortion and slightly tangential distortion. We must therefore deal with the lens distortion. Let $(x,y)$ be the ideal normalized image coordinates, and $(\check{x}, \check{y})$ are the corresponding distorted normalized image coordinates. According to previous works \cite{QCAV_xu:Heikk97,QCAV_xu:WeiM94}, we have
\begin{equation}
\label{distortion}
\begin{cases}
\check{x}=x(1+k_{1i}r^2+k_{2i}r^4)+2p_{1i}xy+p_{2i}(2x^2+r^2)\\
\check{y}=y(1+k_{1i}r^2+k_{2i}r^4)+p_{1i}(2x^2+r^2)+2p_{2i}xy
\end{cases}
\end{equation}
where $r^2=x^2+y^2$, $k_{1i}$ and $k_{2i}$ are the radial distortion coefficients of the $i$-th viewpoint, and $p_{1i}$ and $p_{2i}$ are the tangential distortion coefficients of the $i$-th viewpoint. Then, we can calculate the observed pixel coordinates from Eqn. \eqref{3D2img} and Eqn. \eqref{inMat}

We initially set all the distortion coefficients to zero, and then refine them in the optimization step.

\section{Calibration Algorithm}
There are two steps in our method for parameter calibration. We first calculate the initial values by using a closed form solution, and then a non-linear iterative algorithm is applied to refine the initial values.
\subsection{Closed form solution}
By applying Zhang's calibration method \cite{zhangcalib} to each viewpoint, we can obtain the closed form solution for the intrinsic matrix $\mathbf{A}_i$ and the extrinsic parameters $[\mathbf{R}_i^{w} \quad \mathbf{t}_i^{w}]$. We can compute for the relative extrinsic parameters from the inverse form of Eqn. \eqref{rel2ext}:
\begin{equation}
\label{ext2rel}
\begin{cases}
\mathbf{R}_i=\mathbf{R}_i^{w}\cdot(\mathbf{R}_0^{w})^{-1}\\
\mathbf{t}_i=\mathbf{t}_i^{w}-\mathbf{R}_i\cdot\mathbf{t}_0^{w}
\end{cases}
\end{equation}
Theoretically, the relative extrinsic parameters should be the same for every frame. However, the results vary for different frames in the presence of noise, because the closed form solutions are computed independently for each viewpoint. To obtain reasonable initial values, we calculate relative extrinsic parameters for all captured frames and then choose the median values.

\subsection{Global optimization}
Thus far, we have obtained the intrinsic and extrinsic parameters through a series of linear methods. However, these parameters are not optimal. Also, the linear methods cannot deal with the lens distortion. As suggested in \cite{zhangcalib,multi-camCalib}, nonlinear optimization is needed to refine the linear solutions.

Suppose that the total number of viewpoints is $N$. We captured $T$ frames of a model plane and there are $M$ points on this model plane. Assuming that these image points $\mathbf{m}_{ijk}$ are corrupted by independent and identically distributed noise, the maximum likelihood estimation (MLE) of the intrinsic and extrinsic parameters can be obtained by minimization of the following function:
\begin{equation}
\label{eqn_opt}
\sum\limits_{i=0}^{N-1}\sum\limits_{j=0}^{T-1}\sum\limits_{k=0}^{M-1} ||\mathbf{m}_{ijk}-\hat{\mathbf{m}}(\mathbf{A}_i,\mathbf{K}_i,\mathbf{RT}_i,\mathbf{RT}_{0j}^w,\mathbf{M}_k)||^2,
\end{equation}
where $\hat{\mathbf{m}}(\mathbf{A}_i,\mathbf{K}_i,\mathbf{RT}_i,\mathbf{RT}_{0j}^w,\mathbf{M}_k)$ is the projection of point $\mathbf{M}_k$ in the $j$-th frame of the $i$-th viewpoint, $\mathbf{K}_i$ represents the distortion coefficients of the $i$-th viewpoint [$k_{1i},k_{2i},p_{1i},p_{2i}$], $\mathbf{RT}_i$ represents the relative extrinsic parameters of the $i$-th viewpoint $[\mathbf{R}_i \quad \mathbf{t}_i]$, and $\mathbf{RT}_{0j}^w$ represents the $j$-th frame's extrinsic parameters for the first viewpoint in the global coordinates $[\mathbf{R}_{0j}^w \quad \mathbf{t}_{0j}^w]$. The minimization is performed by using the Levenberg-Marquardt algorithm\cite{QCAV_xu:optimbook}, which is initialized with the linear solution obtained from the closed form.

\begin{figure}[!t]
\centering
\includegraphics[width=3in]{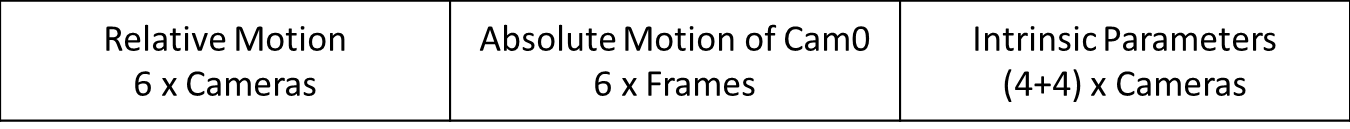}
\caption{Parameters for optimization}
\label{params}
\end{figure}

\begin{figure}[!t]
\centering
\includegraphics[width=3in]{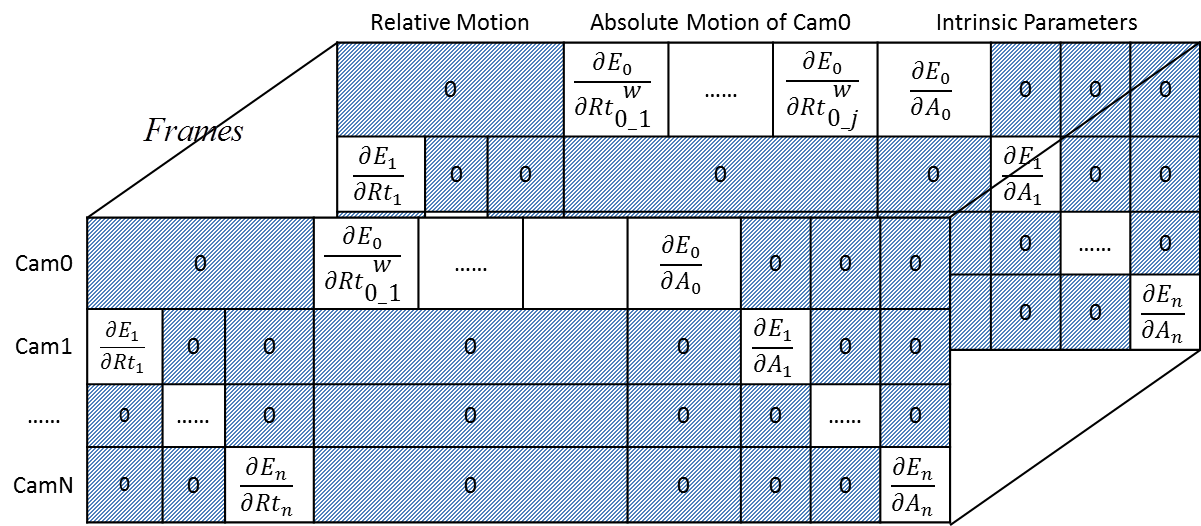}
\caption{Jacobian matrix for optimization}
\label{Jacobian}
\end{figure}

Optimization is carried out for all parameters including the intrinsic matrices, the distortion coefficients and all extrinsic parameters. The components of the parameters for optimization are shown in Fig. \ref{params}. Our implementation is very flexible in that there is no limitation to the number of viewpoints, and the intrinsic parameters are optional for users during runtime.
We assume that the parameters are independent, and Fig. \ref{Jacobian} illustrates the structure of the Jacobian matrix.
The main task of Algorithm 1 is to update the Jacobian matrix and then pass it on to the Levenberg-Marquardt procedure.

\begin{algorithm}[!t]
\caption{Global Optimization}
\begin{algorithmic}[1]
\REPEAT
	\FOR {all frames}
		\FOR {all viewpoints}
			\STATE Compose the absolute motion matrices for each viewpoint
			\STATE Compute the re-projection error
			\STATE Compute the derivation of the intrinsic and extrinsic parameters
			\STATE Calculate the derivation of the relative motion
			\STATE Update the elements of the Jacobian matrix related to relative motion and all intrinsic parameters
		\ENDFOR
		\STATE Update the elements of the Jacobian matrix related to the absolute motion of the first viewpoint
	\ENDFOR
	\STATE Compute the total re-projection error of all viewpoints and frames
	\STATE Launch Levenberg-Marquardt Algorithm to update all parameters
\UNTIL{termination criteria are met}
\end{algorithmic}
\end{algorithm}

\section{Experiments and Application}
The proposed algorithm has been tested on both computer simulated data and experiments with real data.
\subsection{Simulation results}
\begin{figure}[!t]
\centering
\includegraphics[width=3in]{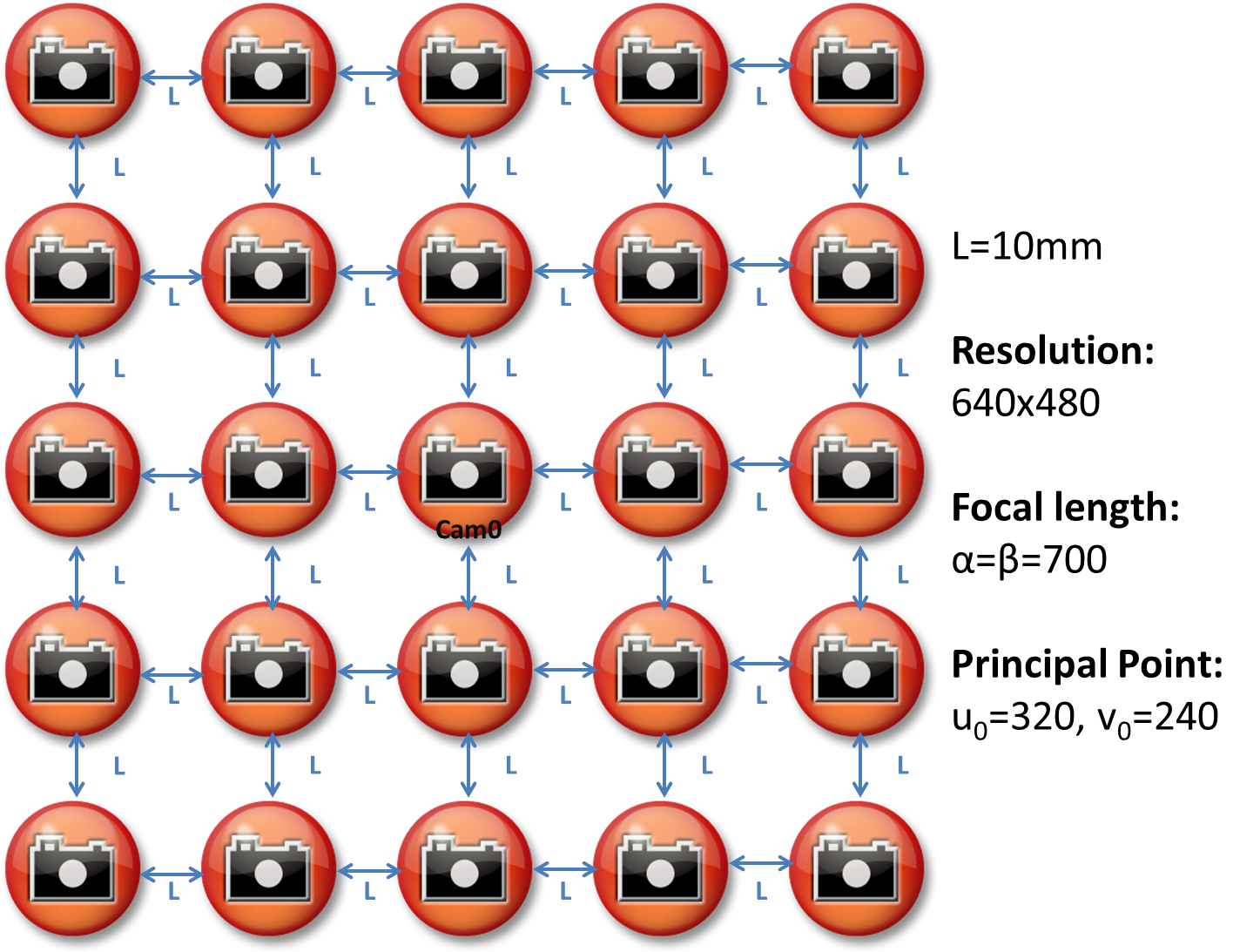}
\caption{Simulation Setting}
\label{sim_setting}
\end{figure}

We performed simulation experiments for a 25 viewpoints (5 horizontal$\times$5 vertical) light field camera. Fig. \ref{sim_setting} shows the configuration of our simulation. The resolution for each viewpoint is $640\times480$. The intervals between the neighboring viewpoints are 10 mm, and we assume all the viewpoints are on the same plane. The simulated light field camera has the following intrinsic parameters for all viewpoints: $\alpha=\beta=700$, $u_0=320, v_0=240$. We simulated 11 frames for the system, each frame is painted with $7\times10=70$ reference points at 20 mm intervals. The distance and orientation of the frames are varied in the simulation. Independent Gaussian noise with 0 mean and $\sigma$ standard deviation (noise level) is added to the simulated image points. The estimated camera parameters are then compared with the ground truth. For each noise level, 100 independent trials are made and average results are shown in the figure.

\begin{figure}[!t]
\centering
\subfloat[Error in focal length]{
\label{fig:subfig_a}
\includegraphics[width=3in]{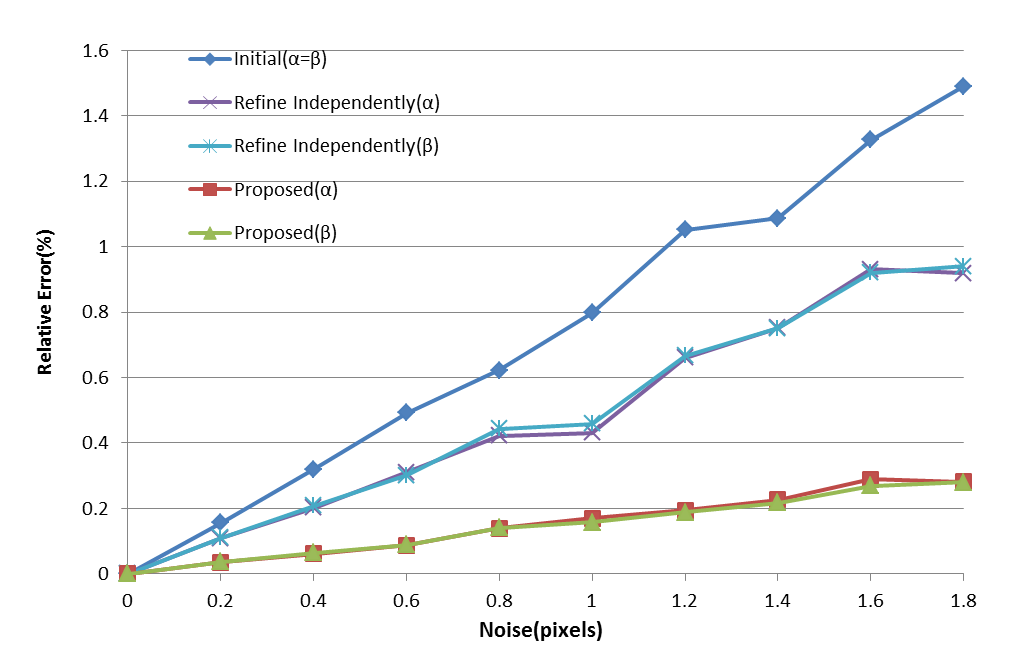}
}
\hspace{80pt}
\subfloat[Error in principal point]{
\label{fig:subfig_b}
\includegraphics[width=3in]{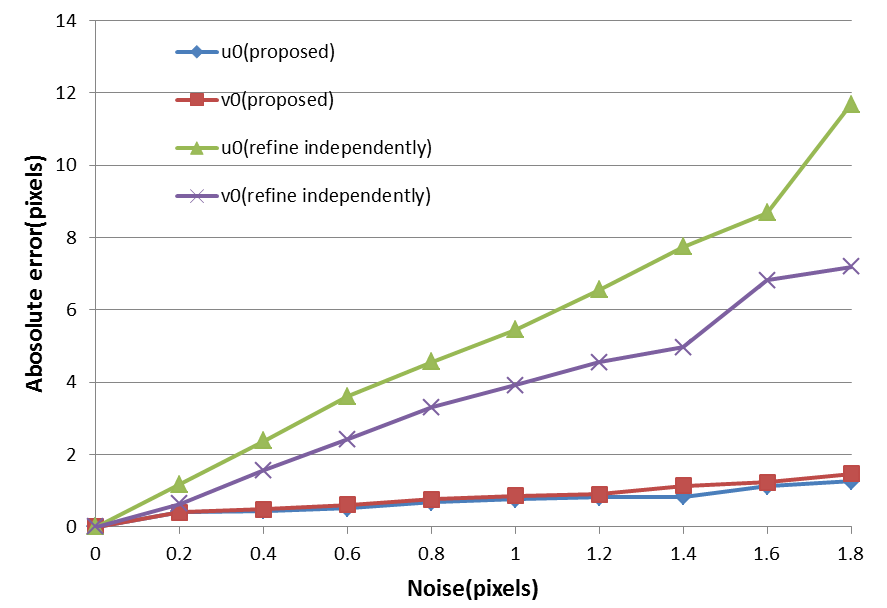}
}
\caption{Error vs. the noise level of the image point of the center camera}
\label{sim_f_pp}
\end{figure}

We measure the relative error for $\alpha$ and $\beta$, and abosulute error for $u_0$ and $v_0$. The noise level is varied from 0.2 pixels to 1.8 pixels. As we can see from Fig. \ref{sim_f_pp}, the errors in the estimated internal parameters of the first camera increase linearly with the noise level. It can be seen that the proposed method yields much better results than other methods. This is because our method is constrained by the geometry relationship between the viewpoints and therefore suffers less from over-fitting to the noise in a single viewpoint.

\begin{figure}[!t]
\centering
\includegraphics[width=3in]{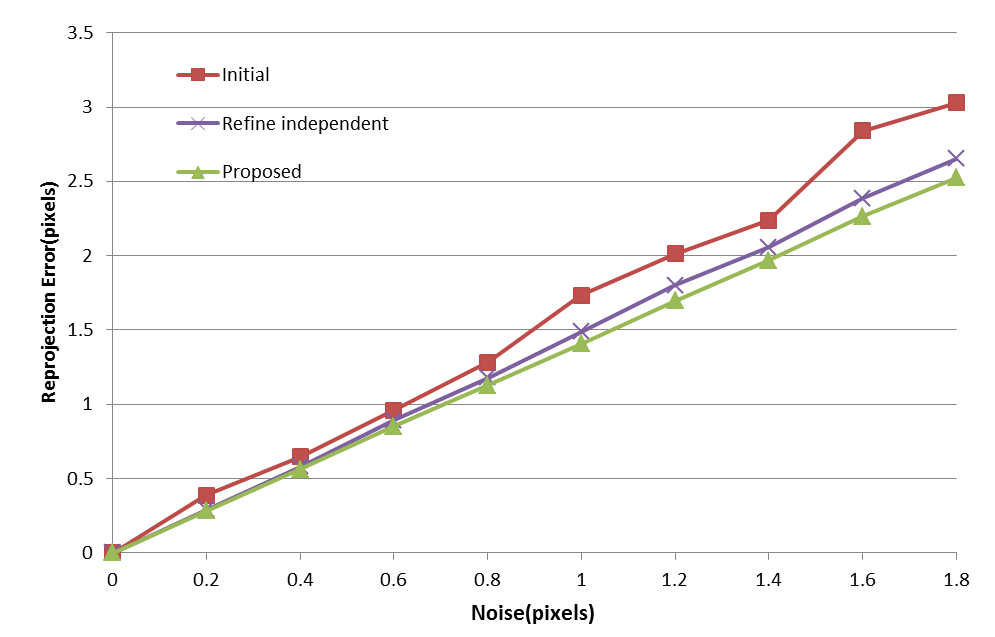}
\caption{Re-projection error vs. the noise level of the image point}
\label{sim_reprojErr}
\end{figure}

Fig. \ref{sim_reprojErr} shows the RMS reprojection error for all the reference points. We can see that our method gets the minimum reprojection error since we globally minimized the reprojection error for all the viewpoints. For the noise level less than 0.6 pixels, the reprojection error can be less than 1 pixel.

\subsection{Results with real data}
The proposed algorithm has also been applied to calibration of the real light field camera. Here, we give an example for calibration of a commercial product, the Pro Fusion25 (ViewPlus Inc., Tokyo, Japan), which has 25 VGA resolution (640$\times$480 pixels) cameras. This camera system can simultaneously capture images from 25 viewpoints (5 horizontal$\times$5 vertical). The central camera in this system is assigned as the first viewpoint.

\begin{figure}[!t]
\centering
\includegraphics[width=3in]{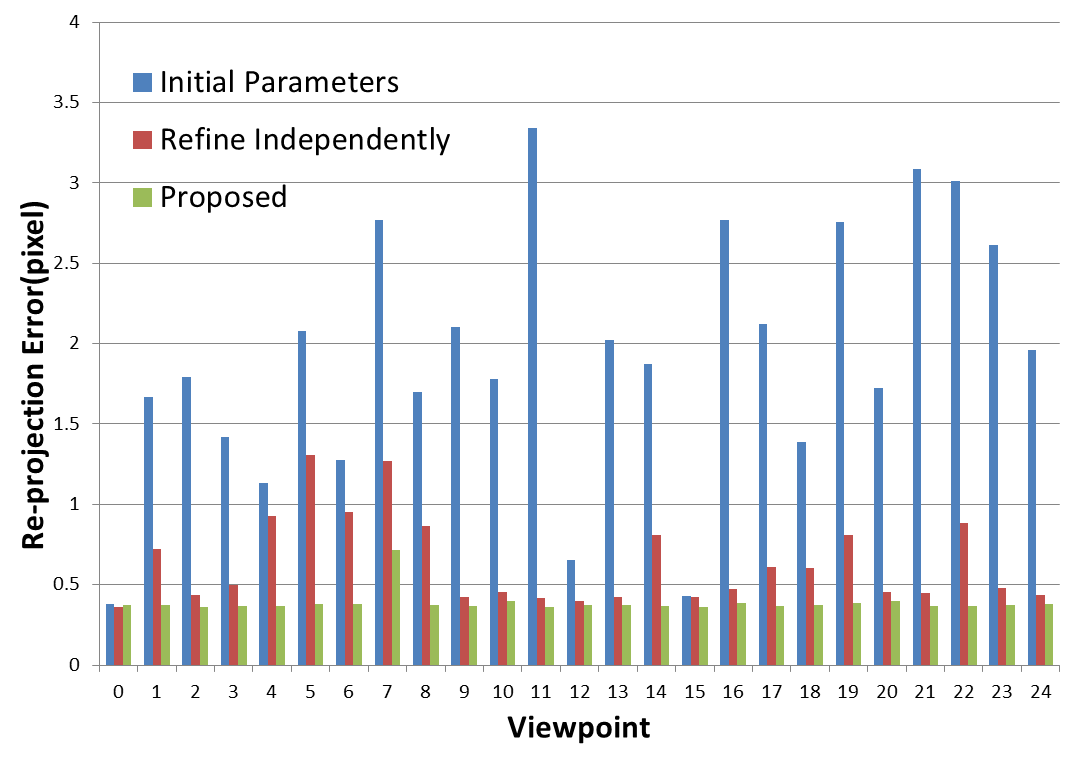}
\caption{Re-projection error for each viewpoint}
\label{CamErrCmp}
\end{figure}

We use the light field camera to capture several checkerboard pattern images. Each checkerboard pattern image contains $7\times10=70$ corner points. We then perform a closed form calibration, refine each viewpoint independently, and finally refine by global optimization. In Table \ref{cmpResult}, we compared the re-projection error of our proposed method with those of another two methods. As shown in the table, we can see that the standard deviation of our proposed method is very small compared to the values for the other two methods. By comparing to the simulation results, it is reasonable that the real data has the re-projection error of 0.4 pixels. The re-projection error for each viewpoint can be seen in Fig. \ref{CamErrCmp}.

When we simply use the initial parameters calculated by the closed form procedure, the re-projection error is more than 2 pixels. When we refine the parameters of each viewpoint independently, and calculate the median values for the relative extrinsic parameters of all captured frames, the total re-projection error becomes less than 1 pixel. However, from Fig. \ref{CamErrCmp}, we can see that the re-projection error for each camera varies. The 5th and 7th viewpoints still have re-projection errors that are larger than 1 pixel. Finally, when all parameters are refined simultaneously, we can obtain a total re-projection error of less than 0.4 pixels. We also confirmed that the re-projection error for each viewpoint is within 1 pixel, i.e. the calibrated image from every viewpoint has no disparity at infinity. However, from Fig. \ref{CamErrCmp}, we can see the 7th viewpoint has a larger re-projection error than the other viewpoints. One reason for this is that its translation along the principal axis is larger than those of the other viewpoints.
 \begin{table}[!t]
 \caption{Comparison of re-projection errors}
 \centering
 \begin{tabular}{|c|c|c|}
 \hline
 Method  & Re-projection error & Standard deviation\\
 \hline
 Initial Parameters  &  2.0672 & 0.7805\\
 \hline
 Refine Independently  &  0.6898 & 0.2670\\
 \hline
 Proposed & 0.3952 & 0.0679\\
 \hline
 \end{tabular}
 \label{cmpResult}
 \end{table}

\subsection{Digital refocusing}
One of the most popular applications of light field cameras is image refocusing after the light field image is captured. However, if the position and orientation of each viewpoint is unknown, we may never obtain a refocused image. Large lens distortion will also affect the results for the refocused image.

\begin{figure}[!t]
\centering
\subfloat[Refocused image with un-calibrated light field]{
\label{fig:subfig_a}
\includegraphics[width=3in]{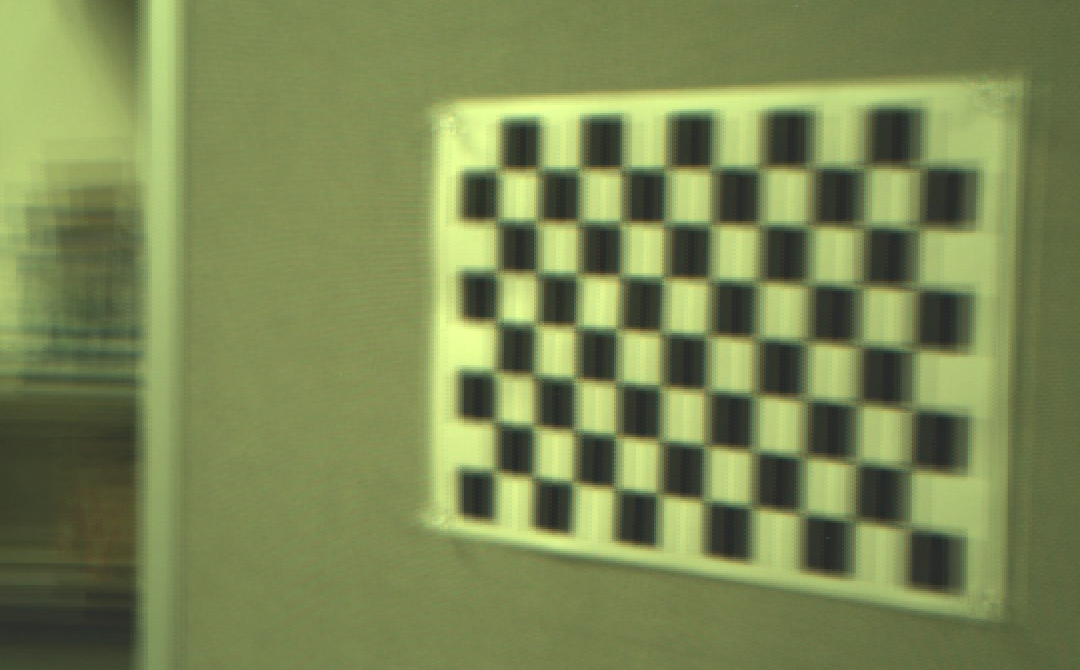}
}
\hspace{80pt}
\subfloat[Refocused image with calibrated light field]{
\label{fig:subfig_b}
\includegraphics[width=3in]{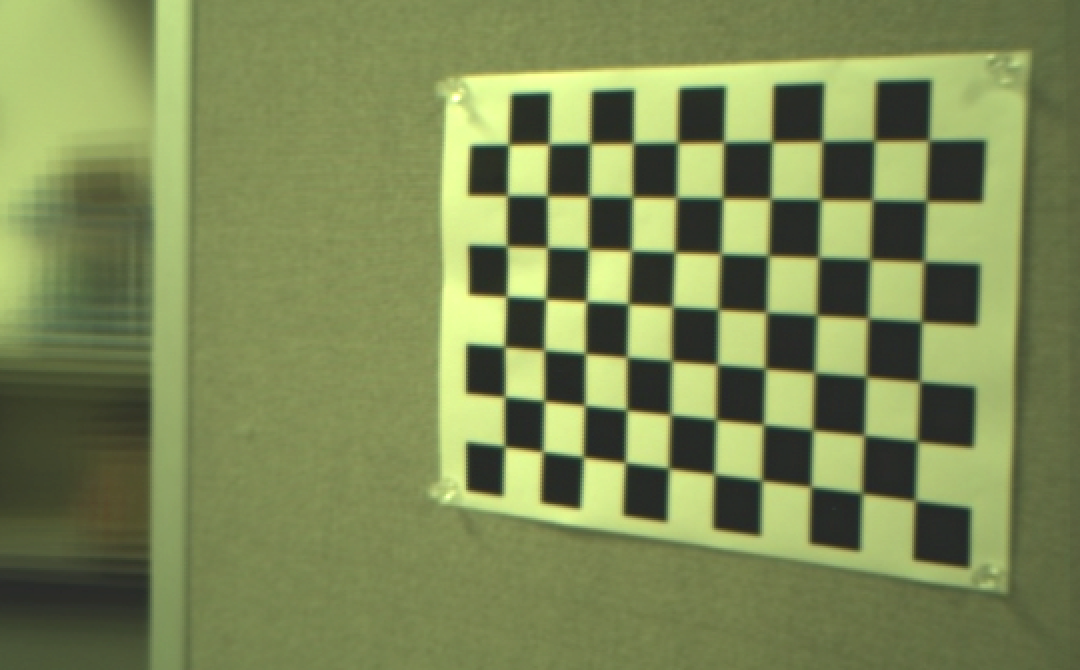}
}
\caption{Refocused image of the calibration chart}
\label{refocusIma}
\end{figure}

We attempted to generate an image focusing on the checkerboard (as shown in Fig. \ref{refocusIma}). The top image is generated from the un-calibrated light field, and we use the positions provided by the official specification. We select the sharpest image visually, but it is still blurred (Fig. \ref{fig:subfig_a}). The bottom image is generated from the calibrated light field, and we can use the positions and orientations obtained from the calibration process. Then, we can perform a warping transform which re-projects the light field image to the target plane, and then simply sum and average the images from all viewpoints. We can see that the image has then clearly focused on the checkerboard (Fig. \ref{fig:subfig_b}).

\begin{figure*}[!t]
\centering
\subfloat[Refocused to the near object]{
\label{fig:subfig_near_un}
\includegraphics[width=2.1in]{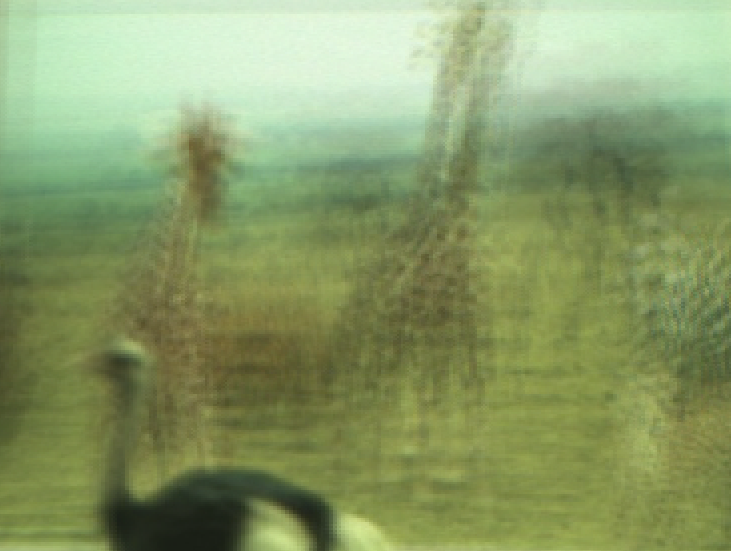}
}
\subfloat[Refocused to the middle object]{
\label{fig:subfig_middle_un}
\includegraphics[width=2.1in]{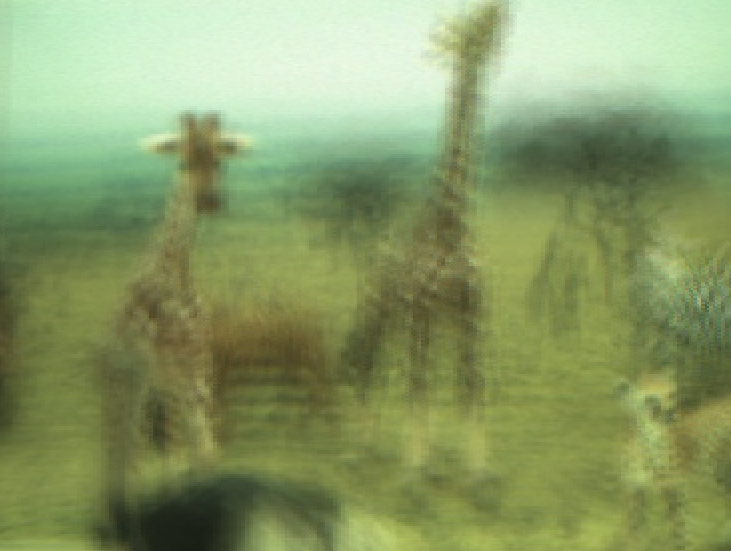}
}
\subfloat[Refocused to the far object]{
\label{fig:subfig_far_un}
\includegraphics[width=2.1in]{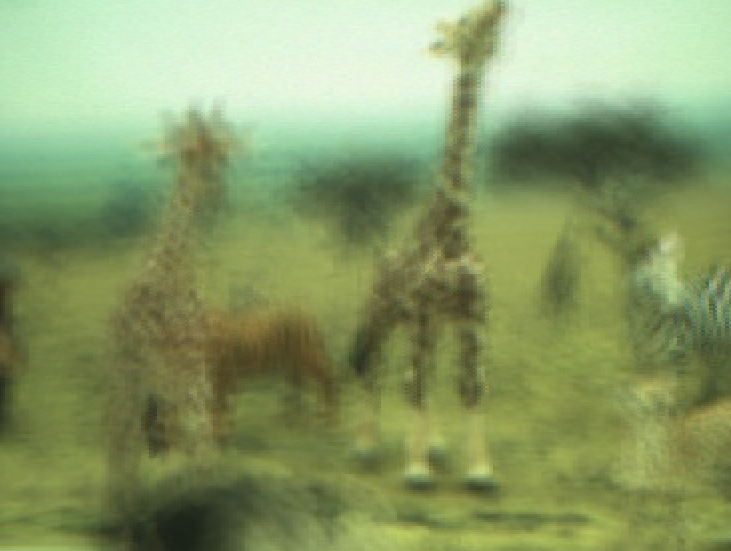}
}
\hspace{80pt}
\subfloat[Refocused to the near object]{
\label{fig:subfig_near}
\includegraphics[width=2.1in]{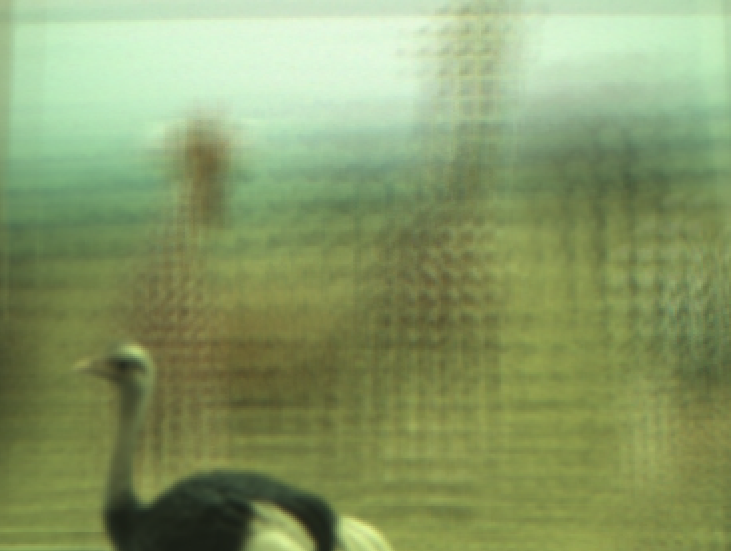}
}
\subfloat[Refocused to the middle object]{
\label{fig:subfig_middle}
\includegraphics[width=2.1in]{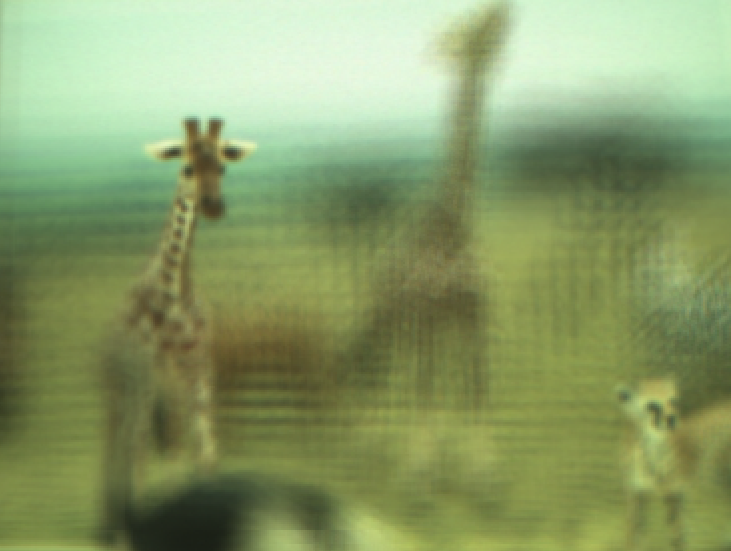}
}
\subfloat[Refocused to the far object]{
\label{fig:subfig_far}
\includegraphics[width=2.1in]{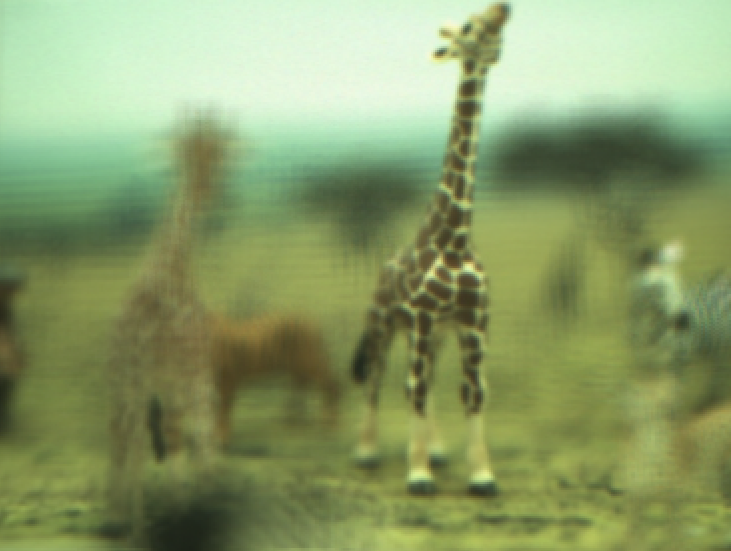}
}
\caption{Refocused image of the real scene with un-rectified(top) and rectified(bottom) light field}
\label{refocusRealScn}
\end{figure*}

We also captured some light fields for the real scene by the commercial product, and then rectify the raw light fields by our proposed algorithm. After that, we can generate images focusing on the different objects (as shown in the bottom row of Fig. \ref{refocusRealScn}). The images from left to right are refocused to the near, middle and far away object respectively. Obviously, the images generated with un-rectified light field (as shown in the top row of Fig. \ref{refocusRealScn}) cannot focus to the object we desired, while the images in bottom row are refocused to the desired object.

\section{Conclusion}
In this paper, we have defined the position-angular representation of a light field, and have performed modeling for a mobile camera array. A calibration algorithm with global optimization has been proposed for the light field camera. The proposed algorithm uses Zhang's plane based style, which is easy to carry out. Our implementation is also flexible; the user can assign the number of viewpoints and the intrinsic parameters are optional for global optimization. Simulation experiments show that our proposed algorithm yield better results than linear solution and refined each viewpoint independently. The proposed method has also been applied to the calibration of a commercial light field camera, and the results show that all intrinsic and extrinsic parameters are optimized, with a total re-projection error of less than 0.4 pixels. We also performed a digital refocusing experiment on the captured light field image. The calibrated light field image can be refocused well to the required target, while this is not possible for the un-calibrated light field image.

The light field camera has been applied in many fields, including image base rendering, synthetic aperture photography and scene geometry reconstruction. If we can extract the camera parameters and perform the warping transformation in real time, then tilt-shift photography can be carried out using the light field camera. Three-dimensional (3D) features can also be detected from the calibrated light field image, so that rotation invariant object recognition in 3D space is also possible with the light field camera.

\bibliographystyle{IEEEtran}
\bibliography{IEEEfull,QCAV_xu}

\end{document}